\documentclass[letterpaper, 10 pt, conference]{ieeeconf}  

\IEEEoverridecommandlockouts                              
\usepackage{multicol}
\usepackage[bookmarks=true]{hyperref}
\usepackage{subcaption}
\usepackage{microtype}
\usepackage{graphics} 
\usepackage{epsfig} 
\usepackage{mathptmx} 
\usepackage{times} 
\usepackage{amsmath} 
\usepackage{amssymb}  
\usepackage{enumerate}
\usepackage{graphicx}
\usepackage{booktabs} 
\usepackage{url}
\usepackage{tikz}
\usepackage{amsfonts}
\usepackage{makecell}
\usepackage{newtxtext, newtxmath}
\DeclareUnicodeCharacter{2212}{-}
\newcommand{\KLD}[2]{D_{\mathrm{KL}}\left( #1 \,\|\, #2 \right)}

\title{\LARGE \bf
Latent Imagination Facilitates Zero-Shot Transfer \\ in Autonomous Racing

}

\author{Axel Brunnbauer$^{1*}$, Luigi Berducci$^{1*}$, Andreas Brandstätter$^{1*}$, Mathias Lechner$^{2}$, \\ Ramin Hasani$^{3a}$, Daniela Rus$^{3}$, Radu Grosu$^{1}$
\thanks{* Indicates authors with equal contributions}
\thanks{$^{1}$ CPS, Technische Universit\"at Wien (TU Wien), Austria}%
\thanks{$^{2}$ Institute of Science and Technology Austria (IST Austria)}%
\thanks{$^{3}$ CSAIL, Massachusetts Institute of Technology (MIT), MA, USA}%
 \thanks{$^{a}$ Correspondence to Ramin Hasani: {\tt\small rhasani@mit.edu}}%
}

\begin{document}

\maketitle
\thispagestyle{empty}
\pagestyle{empty}

\begin{abstract}
World models learn behaviors in a latent imagination space to enhance the sample-efficiency of deep reinforcement learning (RL) algorithms. While learning world models for high-dimensional observations (e.g., pixel inputs) has become practicable on standard RL benchmarks and some games, their effectiveness in real-world robotics applications has not been explored. In this paper, we investigate how such agents generalize to real-world autonomous vehicle control tasks, where advanced model-free deep RL algorithms fail. In particular, we set up a series of time-lap tasks for an F1TENTH racing robot, equipped with a high-dimensional LiDAR sensor, on a set of test tracks with a gradual increase in their complexity. In this continuous-control setting, we show that model-based agents capable of learning in imagination substantially outperform model-free agents with respect to performance, sample efficiency, successful task completion, and generalization. Moreover, we show that the generalization ability of model-based agents strongly depends on the choice of their \emph{observation model}. We provide extensive empirical evidence for the effectiveness of world models provided with long enough memory horizons in sim2real tasks. 
\end{abstract}

\section{Introduction}
Deploying deep reinforcement-learning (RL) agents in the real world is difficult. This is because they require running a significantly large amount of episodes to obtain reasonable performance \cite{sutton2018reinforcement}. This performance is only tractable in simulation environments. Subsequently, the agents should also overcome the challenges of transferring learned dynamics from simulation to the real world. 

In RL settings, it was shown that by learning representations of the state-space model from high-dimensional observations and using them as a predictive model to train policies in imagination, we gain performance, sample efficiency, and robustness~\cite{hafner2019dream}. This world model algorithm~\cite{ha2018worldmodels} was called Dreamer, and it demonstrated great performance in learning long-horizon visual control tasks and Atari games~\cite{hafner2020mastering}. 

\begin{figure}[t]
    \centering
    \includegraphics[width=1\columnwidth]{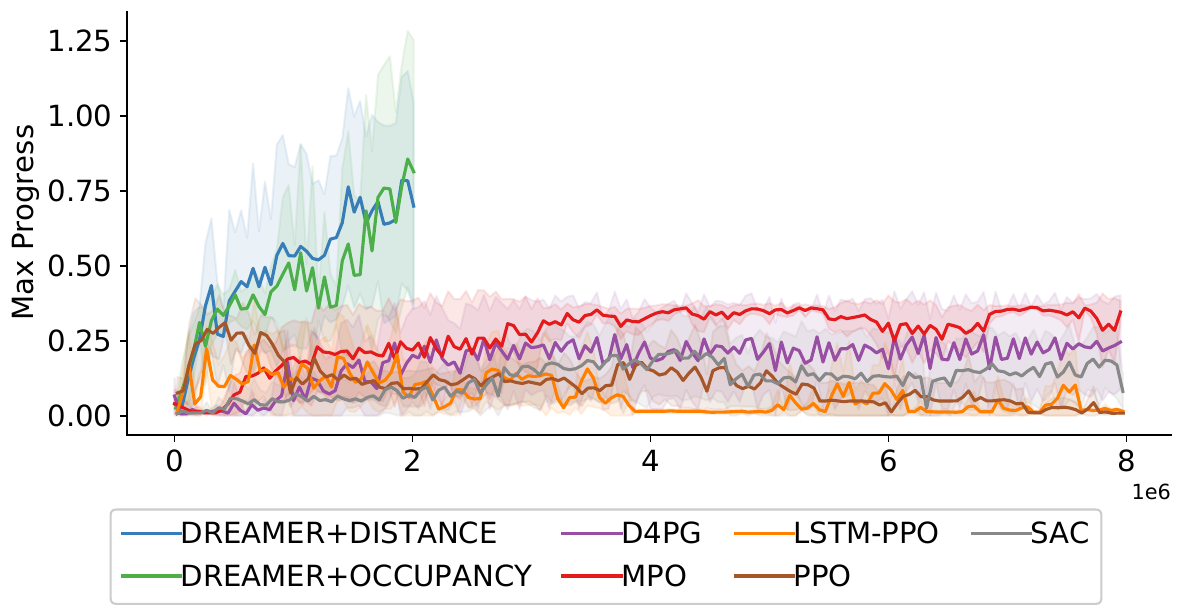}
    \caption{Model-based deep-RL (Dreamer) solves autonomous racing in complex tracks where all advanced model-free methods fail.}
    \label{fig:compare_all_models_austria}
\end{figure}

Providing a model of the world (state transition probability) to the agents (even though in simulation) improves sample efficiency and, in some cases, the performance of a deep RL agent \cite{ha2018worldmodels,hafner2019learning,hafner2019dream,hafner2020mastering}. However, there are still open research questions to be investigated: For instance, how would world model compartments improve the sim2real performance of RL agents in real-world applications? Where is the boundary between the generalization capability of model-based algorithms such as Dreamer, compared to advanced model-free agents such as soft actor-critic \cite{haarnoja2018soft}, distributional models \cite{barthmaron2018d4pg}, and policy gradient \cite{schulman2017proximal,abdolmaleki2018mpo}, in sim2real applications?

In this paper, we aim to provide answers to the above questions. We construct a series of time-lap autonomous racing tasks with varying degrees of complexity for an F1TENTH robot. The task is to learn to drive autonomously directly from high-dimensional LiDAR inputs to successfully finish a lap without collisions. 

We observe that as the complexity of the map increases, model-free agents get stuck in similar local minima and cannot learn to complete the task. Contrarily, Dreamer can learn a proper state transition model, and given enough imagination horizon, solves time-lap tasks of arbitrary complexity (See Fig. \ref{fig:compare_all_models_austria}). Moreover, we discover that model-based techniques demonstrate desirable transferability: Dreamer agents trained on a single map can generalize well to test tracks they have never seen before. In summary, our \textbf{main contributions} are as follows:

\begin{itemize}
\setlength{\itemsep}{0pt}
    \item We demonstrate the effectiveness of advanced model-based deep RL compared to model-free agents in the real-world application of autonomous racing.
    \item We show the transferability of advanced model-based deep RL agents to the real-world applications where model-free agents fail.
    \item We empirically show that the learning performance and generalization ability of Dreamer in sim2real applications highly depends on the choice of the observation model, its resulting state transition probability model, and its imagination horizon.
\end{itemize}

\section{Problem Definition}
\label{section:problem_definition}
We set out to design deep RL models that are able to learn to autonomously complete time-lap racing tasks. Considering the real-world conditions, we formalize the problem as a Partially-Observable Markov Decision Process (POMDP).

\noindent \textbf{Notation and terminology.} A POMDP is defined as a tuple $(S, A, \Omega, \mathcal{O}, \mathcal{T}, \mathcal{R})$, where $S, A, \Omega$ are the sets of states, actions and observations, respectively. $\mathcal{O}$ and $\mathcal{T}$ denote stochastic observation and transition functions, and $\mathcal{R}$ is a deterministic reward function. The transition function $\mathcal{T}$ models the system dynamics and includes its uncertainty. It is defined as a stochastic function $\mathcal{T} : S \times A \times S \rightarrow [0, 1]$ which returns the transition probability between two states by applying actions. The observation function $\mathcal{O}$ models the system's perception, including its uncertainty. Its stochastic formulation $\mathcal{O} : S \times \Omega \rightarrow [0, 1]$ returns the probability of perceiving an observation in a given state. The reward function is deterministic $\mathcal{R} : S \times A \times S \rightarrow \mathbb{R}$, which returns the credit assigned to a transition. We discuss our reward shaping subsequently. 

\begin{figure}[t]
\begin{center}
\centerline{\includegraphics[width=0.7\columnwidth]{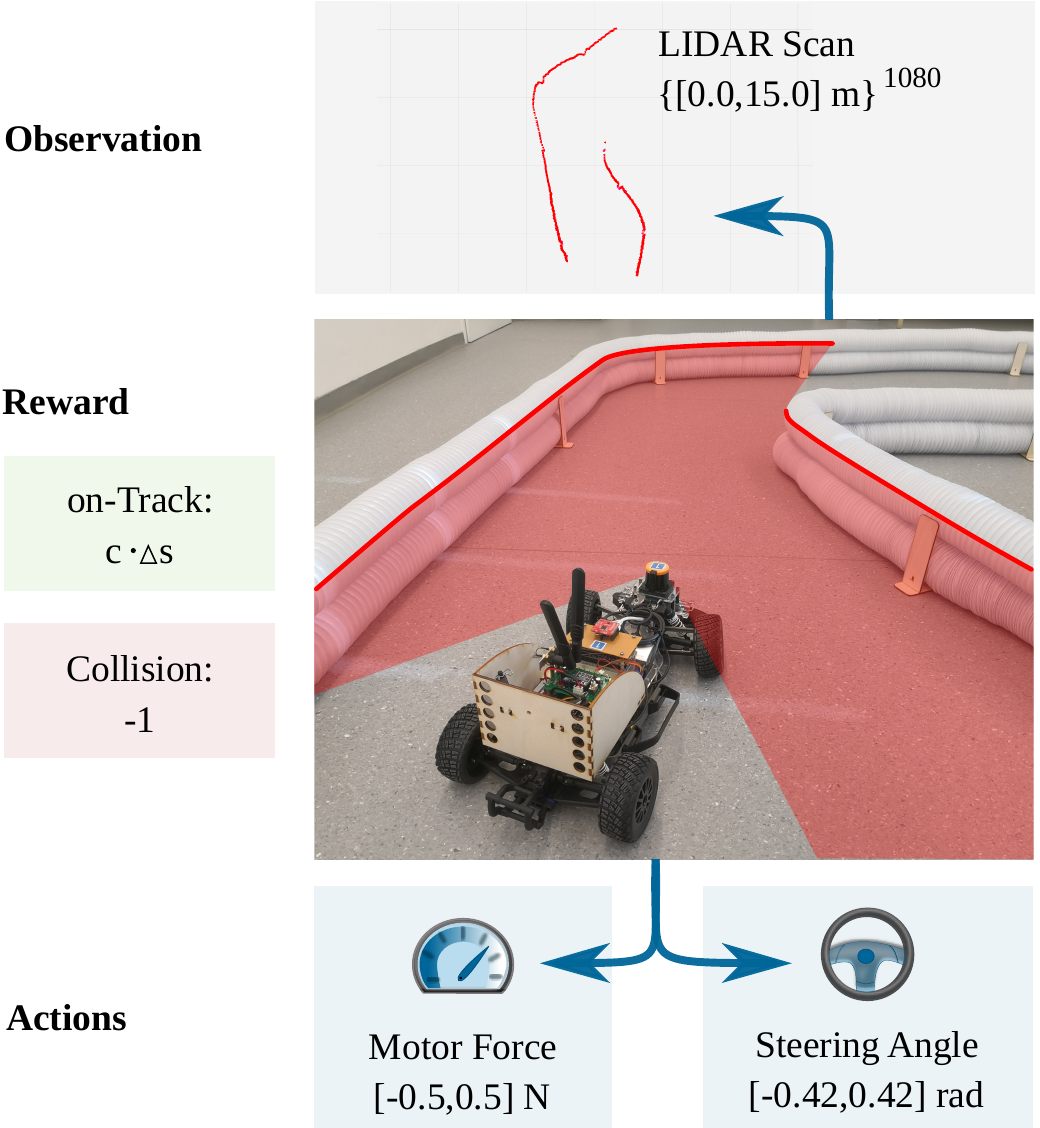}}
\caption{Racing-agent setup: observations, actions and reward.}
\label{figure:mdp_racecar}
\end{center}
\end{figure}

\noindent \textbf{Racing-agent setup.} In autonomous racing, the car drives along a race track by controlling its actuators with continuous actions $\boldsymbol{a}=(F, \alpha)^T$, where $F$ is the motor force applied to reach a fixed target velocity, and $\alpha$ is the steering angle.
In this scenario, the observations are range measurements obtained by a LiDAR sensor with 1080 range measurements evenly distributed over a 270° field of view (see Fig. \ref{figure:mdp_racecar}).

\noindent \textbf{How do we aim to solve this racing objective?} 
Traditional control techniques model the state space as a set of continuous variables such as the car's pose, velocity, and acceleration. These quantities are usually estimated from observations by dedicated perception and filtering modules. State-space planning commonly uses sophisticated dynamics models for cars. This demands non-trivial system identification techniques. Contrarily, we aim to replace the entirety of such modules by learning an abstract state representation and transition model in a self-supervised manner by building upon recent advances of model-based RL \cite{hafner2019dream}.

\section{Related Work}
\noindent We include works that are closely related to our contributions.  

\noindent \textbf{Model-free RL in robot control.}
Robot control has been a playground for RL algorithms for decades \cite{sutton2018reinforcement,hasani2020natural}. Model-free algorithms \cite{schulman2017proximal,haarnoja2018soft,barthmaron2018d4pg} achieved state-of-the-art results in various continuous control tasks, but their inherent sample inefficiency \cite{yang2018sampleeff} makes it challenging to apply them in real-world settings. Off-policy \cite{mnih2013dqn,haarnoja2018soft} methods reduce the sample complexity by reusing old experience to some degree. However, they might induce errors when deployed in a closed-loop with the real-world environment. Thus, model-free approaches are often trained in simulation before being deployed on real robots. This leads to difficulties when reproducing learned behaviors in the real setting \cite{Zhu2020robotics,sim2real_survey_2020}.
To overcome simulation mismatches, some approaches rely on domain randomization and subsequent fine-tuning. In order to learn more robust policies \cite{peng2018sim2real}, other works refined the physics simulator \cite{sim2real_quadruped_robot_2018,kaspar2020sim2real} or directly trained the policy on real-world robots \cite{Haarnoja2019learningToWalk,Levine_EndToEnd_withoutRewardEngineering}.

\noindent \textbf{Model-based RL in robot control.} Model-based approa\-ches, on the other hand, can leverage their learned dynamics model by either planning or generating new training experience \cite{sutton1991dyna}. Problems arise when no sufficiently accurate dynamics model can be learned from data. 
An early representative of model-based RL algorithms is PILCO \cite{deisenroth2011pilco} which learned a Gaussian Process from the system's states. 
Recent works on world models \cite{ha2018worldmodels,hafner2019learning,hafner2019dream, schwarting2021deep} proved the feasibility of learning a  dynamics model for POMDP's by using noisy high-dimensional observations instead of accurate states.
They achieve better than state-of-the-art results on various simulated control tasks while being significantly more sample efficient.
However, to the best of our knowledge, their application and challenges in real-world testbeds have been less attended, which we will explore in the present work. 

\noindent \textbf{Optimal control for autonomous racing.}
Autonomous racing has been an active field of research in the control community. Traditional control does not solve the problem end-to-end but divides it into independent sub-problems: perception, planning, and control. Recent success in the context of Formula Student \cite{kabzan2019amz,kabzan2019model,andresen2020mappingplanning_autonomousracing} has been achieved by Model Predictive Control (MPC) and by engineering the perception pipeline using sensor fusion of LiDAR and RGB cameras. Several approaches plan an optimal trajectory \cite{velenis1528936,rucco7047758,timings2013minimum,vazquez2020hierarchical_motion_planning}. However, these techniques usually require detailed global information (e.g., map) and an accurate dynamics model, not available in a POMDP setting, as in our case. Here, we use a fully automated process that learns an approximated dynamics model in an unsupervised fashion by only having access to LiDAR observations.

\noindent \textbf{RL for autonomous racing.}
A large body of research made use of camera images and adopt model-free methods \cite{jaritz_EndToEndRaceDrivingWithDeepRL, riedmiller_LearningToDriveARealCarIn20Minutes, kendall_LearningToDriveInADay}. In \cite{fuchs2020super}, the authors used SAC to control a racing car in simulation by feeding the controller with state information and ad-hoc features about the road curvature.
To deal with the complexity of continuous action spaces, a common technique is to discretize the domain, but this approach is not scalable. 
Another recurrent trend consists of combining MPC and deep RL \cite{bellegarda2020mpc_drl_trajplanning,williams2017mpc}. However, MPC requires to efficiently perform sampling through the model to scale with respect to the time constraints. Conversely, the adoption of a policy network is more efficient and suitable for an online setting.

\noindent \textbf{Imitation Learning.} Imitation learning is the process of learning observation-to-action mappings from supervised data \cite{schaal1999imitation}. It allows for behavior cloning by data aggregation (Dagger) \cite{ross2011reduction} via supervised learning modes \cite{vorbach2021causal,lechner2020gershgorin} or by inverse RL \cite{ng2000algorithms}. Imitation learning has been further adapted to imperfect \cite{wu2019imitation} and incomplete demonstrations \cite{sun2019adversarial,lechner2021adversarial}. In the context of end-to-end control of autonomous vehicles, learning from expert demonstrations is the dominant choice of algorithm \cite{lechner2020neural,lechner2019designing}, because other RL methods would require efficient simulation platforms to achieve desirable performance \cite{amini2020learning}. For autonomous racing, as we already have a sample efficient simulator available, we settle to use model-based RL algorithms over imitation learning. 

\section{Adapting Dreamer for Autonomous Racing}

In this section, we revisit Dreamer~\cite{hafner2019dream} and discuss its merits for autonomous racing. We present two observation models that we utilize to learn the latent-dynamics model. We then formulate a specialized reward function.

\noindent\textbf{Introduction.} Dreamer is a model-based deep-RL algorithm that has recently achieved state-of-the-art performance in simulated standard RL benchmarks~\cite{hafner2019dream}. 
Dreamer learns a model of the system dynamics from high-dimensional observations in an unsupervised fashion and uses it to produce latent-state sequences to train an agent using an actor-critic algorithm \cite{hafner2019dream}.
Dreamer implements the dynamics model as a recurrent state-space model (RSSM) with both stochastic and deterministic components \cite{hafner2019learning}.
It consists of four components, which are all implemented as deep neural networks~\cite{hafner2019dream}:
\begin{align}
    &\text{Representation model:}       &p_\theta(s_{t} | s_{t − 1}, a_{t − 1}, o_{t}) \\
    &\text{Observation model:}  &q_\theta(o_{t} | s_t) \\
    &\text{Reward model:}     &q_\theta(r_{t} | s_t)\\
  &\text{Transition model:}     &q_\theta(s_{t} | s_{t-1}, a_{t-1})
\end{align}

%
%
%
%
%

All components are jointly optimized to maximize the variational lower-bound as follows, where  $\KLD{P}{Q}$ is the Kullback–Leibler divergence of distributions $P$ and $Q$:
\begin{align}
    &\mathcal{J}^{t}_{\mathcal{O}} = \ln q(o_{t} | s_{t})~~~~~~\mathcal{J}^{t}_{\mathcal{R}} = \ln q(r_{t} | s_{t}) \nonumber \\
    &\mathcal{J}^{t}_{\mathcal{D}} = -\beta \KLD{p(s_t|s_{t-1}, a_{t-1}, o_t)}{q(s_t|s_{t-1}, a_{t-1})}  \nonumber\\
    &\mathcal{J_{REC}} = \mathbb{E}_p \left[ \sum_{t} \mathcal{J}^{t}_{\mathcal{O}} + \mathcal{J}^{t}_{\mathcal{R}} + \mathcal{J}^{t}_{\mathcal{D}} \right]
\end{align}

In contrast to~\cite{hafner2019dream}, our representation model uses a multilayer perceptron (MLP) to encode the observations instead of convolutional layers. The reason for this design decision is that LiDAR scans are not as high-dimensional as RGB images, and further compression does not show better performance in our empirical evaluation.

\begin{figure}
    \centering
    \includegraphics[width=\columnwidth]{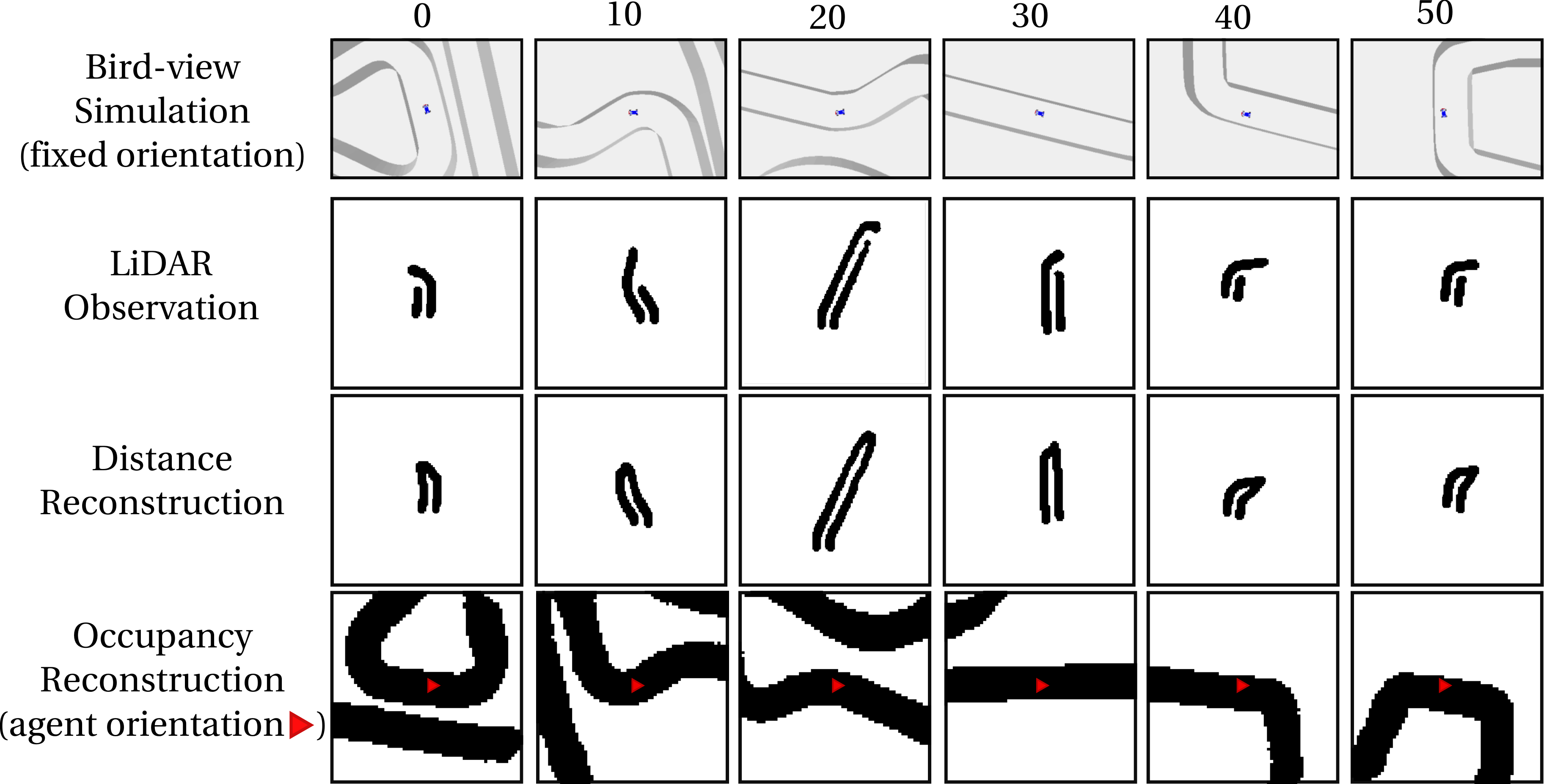}
    \caption{Observation models and their reconstruction methods. Observations are in agent coordinates. \textbf{Row 1:} bird-view of the racecar in simulation, \textbf{Row 2:} LiDAR scan in racecar coordinates, \textbf{Row 3:} reconstructed LiDAR scan, \textbf{Row 4:} reconstructed local occupancy map}
    \label{fig:reconstruction}
\end{figure}

\noindent \textbf{Observation model.} This model reconstructs observations from latent states and is used only in the training phase to generate learning signals for the representation model. In this work, we propose two observation models for autonomous racing. One simply reconstructs the LiDAR scan, which is analogous to the original model, and the other is based on reconstructing occupancy maps. In Figure~\ref{fig:reconstruction} we show snapshots of the simulation, the associated LiDAR observation, and reconstructions sampled from these models.

The \emph{Distance reconstruction} model, implemented by MLPs, predicts a Gaussian distribution over the distance measurement of each LiDAR ray from latent states. We call it Dreamer-Distance. Here, the observation is both the representation model's input and the observation model's output.

Our novel \emph{Occupancy reconstruction} model attempts to predict an occupancy grid map of the agent's surrounding based on its current state.
Specifically, the model generates the parameters for a multivariate Bernoulli distribution which, for every pixel, models the probability of being occupied.
The observation model is implemented using transposed convolutions to construct a 2D grayscale image and is trained by providing patches of the true occupancy grid map.
We call the resulting agent  Dreamer-Occupancy. 

\noindent\textbf{Actor-critic on latent imagination.}
Having first learned a world model from observations, Dreamer then learns a policy purely on latent state-action sequences~\cite{hafner2019dream}.
This approach allows Dreamer to efficiently produce thousands of training sequences for RL without direct environment interaction, which results in more data-efficient learning. Starting from these imagined sequences, Dreamer uses an actor-critic algorithm to train the agent.
The action model (policy) aims to predict the best action while the value model estimates the value for each latent state:
\begin{align}
    &\text{Action model:}       &q_\phi(a_{t} | s_{t}) \\
    &\text{Value model:}        &q_\psi(v_{t} | s_{t})
\end{align}
To trade off bias and variance in value estimation, Dreamer uses an exponentially weighted sum over different horizons~\cite{hafner2019dream}.
Moreover, the training process leverages the availability of a learned dynamics model by back-propagating through it. This leads to more effective gradient updates.
%

\begin{figure}[t]
\begin{center}
\centerline{\includegraphics[width=0.45\textwidth]{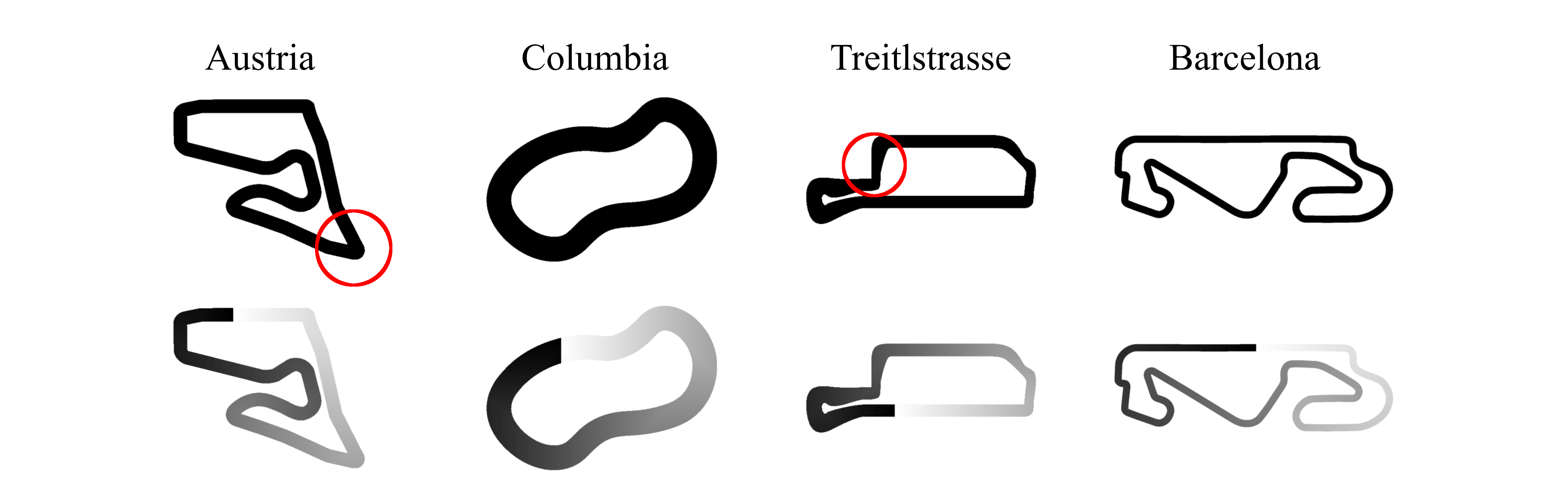}}
\caption{Top-view of the race tracks and relative Progress Maps adopted for reward design. Austria (length = $79.45m$), Columbia ($61.2m$), Treitlstrasse ($51.65m$) are used for training, Barcelona ($201m$) is used only for evaluation. Red circles indicate difficult parts of the track. The red circles show where all model-free agents fail to improve their performance.}
\label{figure:racecar_maps}
\end{center}
\end{figure}

\noindent\textbf{Reward shaping.} The reward design is critical for learning a policy. Sparse rewards over long episodes makes the learning problem challenging. Starting from a simple approach that rewards agents only when finished the race, we gradually refine the reward to provide a high-density learning signal.
\noindent The reward signal we propose for this task is defined as
\begin{equation}
    c * |p_{t}-p_{t-1}| = c * \Delta p_t,
\label{eq:progress}
\end{equation}
where $p_t$ denotes the progress that has been made on the track at time $t$ and $c$ is a constant scalar.
When colliding with the wall or other objects, the agent receives a penalty and the episode terminates (see also Figure~\ref{figure:mdp_racecar}).
The progress is computed by a distance transform applied to the gridmaps for each track. This yields a normalized progress estimate for each pixel on the map at a resolution of ~5cm per pixel. Therefore, the progress value can be easily queried from these maps given the current pose of the car.
In Figure~\ref{figure:racecar_maps} (bottom) we show the precomputed progress maps for each of our tracks. The progress value ranges from 0 (light) to 1 (dark), where a progress of 1 corresponds to one full lap.



\section{Experiments}
Here, we describe our experiments and evaluate the performance of model-free versus model-based RL algorithms.

\subsection{Experimental Setup}
\label{subsection:experimental_setup}
For our experiments, we train the agents in simulation before we transfer them to real cars. In the following, we provide an overview of the simulation and real-world setup as well as the training process.

\noindent \textbf{Simulation environment.} To simulate a car model, we use the open-source physics engine PyBullet~\cite{pybullet}. 
The car model is a rigid body system based on the URDF model in~\cite{babu2020f1tenth}.
Our training environment can simulate a broad set of sensory inputs, such as LiDAR sensors, RGB cameras, and odometry. 
The model is actuated by applying force to the steering and acceleration joints. 
%

\noindent \textbf{Training pipeline.} During training in simulation, we place the agent randomly on the track at the beginning of an episode. Each training episode has a maximum length of $2\,000$ timesteps, resulting in 20 seconds of real-time experience. Agents learn to directly maximize the progress covered in a small, predefined time interval. 
Observations and actions are normalized. To account for latency experienced during testing and to increase the effectiveness of actions, we repeat each action multiple times. We regularly evaluate the agent by placing it on a fixed starting position for each track and let it run for at most $4\,000$ timesteps (i.e., 40 seconds) and average the maximum progress reached over five consecutive trials. Dreamer is trained for 2 million timesteps. The model-free baselines are trained for 8 million timesteps.

\noindent \textbf{Hardware setup.} 
The hardware platform is the F1TENTH race series \cite{o2020tunercar}. It consists of an off-the-shelf model race car chassis, with a Traxxas Velineon 3351R brushless DC electric motor, which is driven by a VESC 6 MkIV electronic speed controller (ESC). The laser range measurements are produced by a Hokuyo UST-10LX LiDAR sensor. On-board computation and control tasks are performed on an NVIDIA Jetson TX2. 
The on-board system runs Ubuntu 18.04 as the base operating system and hosts the core services for the Robot Operating System (ROS) stack~\cite{ros}.
To run Dreamer agents on our hardware, we use Docker. 
A ROS interface node is used to translate observation and action messages. The motor force commands are processed by integration to get the desired speed values. We added an adaptive low pass filter for the steering commands to protect the servo from high frequent steering operations.

\begin{figure*}[t]
     \centering
     \begin{subfigure}[b]{0.49\textwidth}
         \centering
         \includegraphics[width=\textwidth]{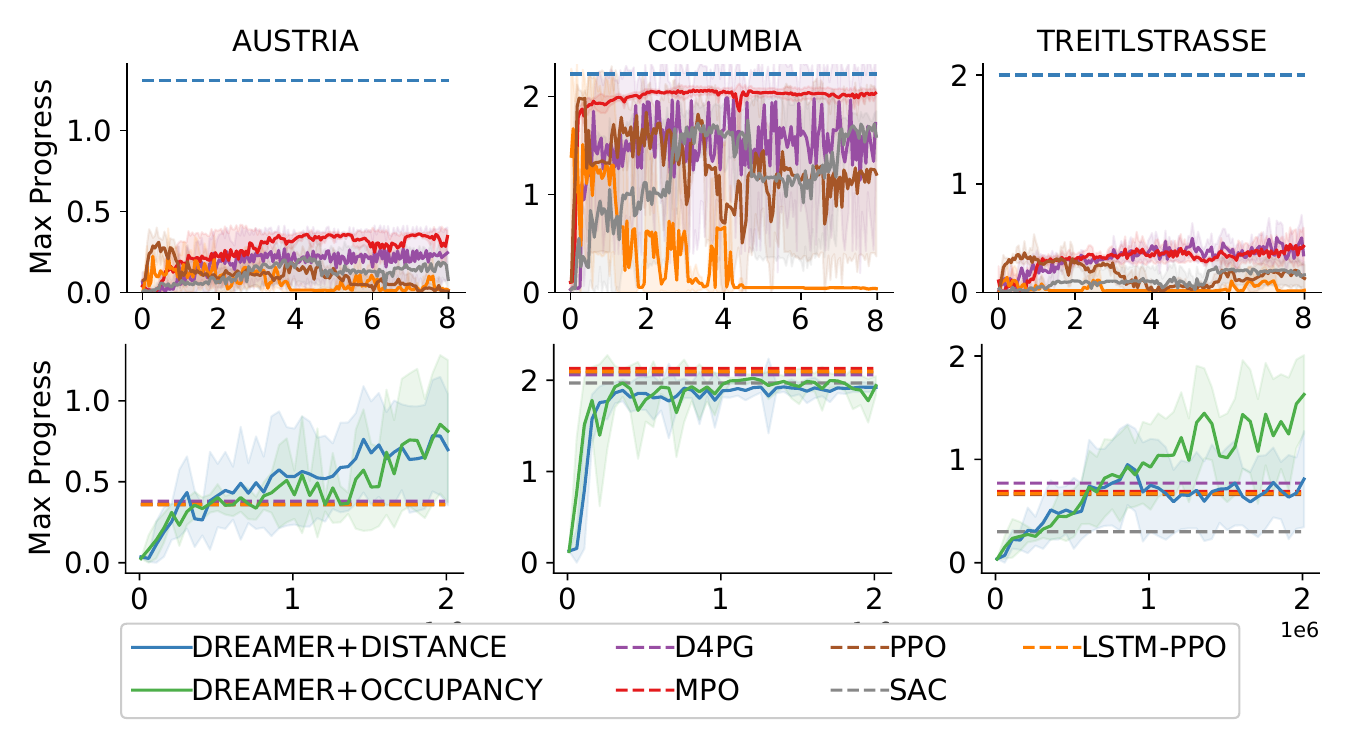}
         \label{fig:learning_curves}
     \end{subfigure}
     \hfill
     \begin{subfigure}[b]{0.49\textwidth}
         \centering
         \includegraphics[width=\textwidth]{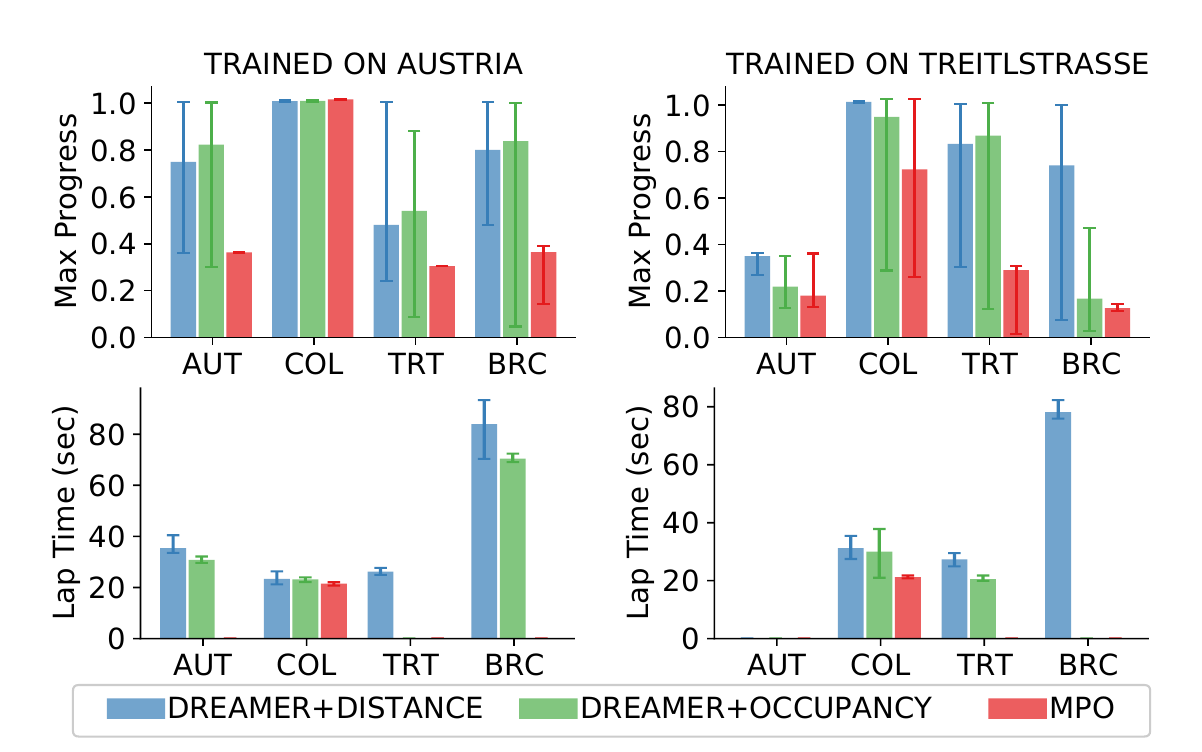}
         \label{fig:progress_transfer}
     \end{subfigure}
     \hfill
    \caption{\textbf{Left:} Learning curves of model-free methods (\textit{top row}) over $8M$ steps and Dreamers (\textit{bottom row}) over $2M$ steps. The dashed lines report the maximum performance obtained by the other algorithms as baselines. Performance averages over 5 runs.
    \textbf{Right:} Maximum progress and lap time of trained models over different tracks in simulation. The bars show the result averaged over 10 episodes on each track. The delimiters show the minimum and maximum achieved. For Lap-Time results, we consider the best episode that finished one full lap.}
    \label{fig:learning_curves_and_transfer}
\end{figure*}

\noindent \textbf{Track difficulty.} To compare the difficulty of the tracks, we classified them according to several characteristics (as also discussed in \cite{BRAGHIN20081503} and \cite{Lamiraux2001Smoothmotionplanning}). In particular, 
we measure the minimal track width, the track length (shortest path), and its minimum curve radius. 
For curves, we measure the radius of the largest circle that tangentially touches the outside of the curve and also touches the inside of the curve. 
In Table~\ref{track-characteristics-table} we summarize the characteristics of the tracks. 

\begin{table}[t]
\caption{Track characteristics.}
\label{track-characteristics-table}
\begin{center}
\begin{small}
\begin{tabular}{lrrr}
\toprule
\textbf{Track}  & \textbf{Min. Width} & \textbf{Length} & \textbf{Min. Radius} \\
\midrule
Austria & 1.86m & 79.45m & \textbf{2.78m} \\ 
Columbia & 3.53m & 61.20m & 7.68m \\ 
Treitlstrasse & 0.89m & 51.65m  & 3.55m \\ 
Barcelona & 1.86m & 201.00m & 2.98m \\ 
\bottomrule
\end{tabular}
\end{small}
\end{center}
\end{table}

\subsection{Experimental Evaluation}\label{section:results}
In this section, we present the baseline algorithms and discuss their performance in various racing scenarios. 
We conduct three different experiments: I) Evaluation of the learning curves of Dreamer and model-free algorithms in simulation, II) Evaluation of the generalization ability by testing the trained models in simulation, and III) Evaluation of the transferability on our real testing platform.

\noindent \textbf{Model-free baselines.} We compare the performance of the Dreamer agent against the following advanced model-free baselines: D4PG, an enhanced version of DDPG~\cite{barthmaron2018d4pg}, MPO, a stable off-policy algorithm~\cite{abdolmaleki2018mpo}, SAC, an off-policy actor-critic algorithm with less sensitivity to hyperparameters~\cite{Haarnoja2019learningToWalk}, 
PPO, an on-policy algorithm~\cite{schulman2017proximal}, and PPO-LSTM, the recurrent version of PPO using long short-term memory~\cite{hochreiter1997long}. PPO-LSTM is chosen as a baseline since policies built by recurrent networks demonstrate remarkable performance in learning to control \cite{hasani2021liquid,hasani2019response,lechner2020learning,hasani2021closed}. We tuned the hyperparameters for each baseline algorithm with Optuna~\cite{akiba2019optuna}.

\noindent \textbf{Learning performance and sample efficiency.}
In Figure~\ref{fig:learning_curves_and_transfer} (left) we show the learning curves of the model-free and model-based algorithms in three different tracks. Except for the simple track Columbia, all advanced model-free methods fail to perform one lap in more complex maps successfully. On the other hand, Dreamer efficiently learns to complete the tasks regardless of the degrees of complexity of the tracks.

As shown in Figure~\ref{fig:learning_curves_and_transfer}, the performance of the Dreamer agents equipped with distance and occupancy reconstruction models are comparable in Austria and Columbia. However, the experiments on Treitlstrasse show better performance with occupancy-map reconstruction, with which the agent can almost complete two full laps over the evaluation-time window. This experiment suggests that the occupancy-map reconstruction speeds up the training process. However, it biases the learning process on the track on which it was trained. This affects the generalization capabilities of the policy, as illustrated in the following experiment.

\noindent \textbf{Performance on unseen tracks.}
In this experiment, we evaluate the cross-track generalizability of the learned policies and demonstrate the domain-adaptation skills of Dreamer. 
We trained polices on a single, fixed track, i.e., Austria (AUT) and Treitlstrasse (TRT) and reported their evaluation performance on other unseen tracks: Columbia (COL) and Barcelona (BRC). We compare Dreamer to the best-performing model-free policy, MPO, and to a Follow-the-gap (FTG) agent, which is a LiDAR-based reactive method that was successfully applied in past F1TENTH competitions~\cite{ftg2012}.

\begin{figure}[b]
    \centering
    \includegraphics[width=0.6\columnwidth]{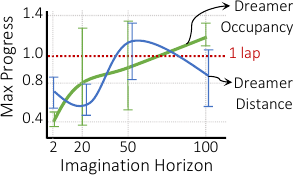}
    \caption{Dreamer's performance vs. imagination horizon. Batch length =$50$, and action repeat = $4$. n=5.}
    \label{fig:compare_horizon}
\end{figure}

\begin{figure*}[t!]
    \centering
    \includegraphics[width=1.4\columnwidth]{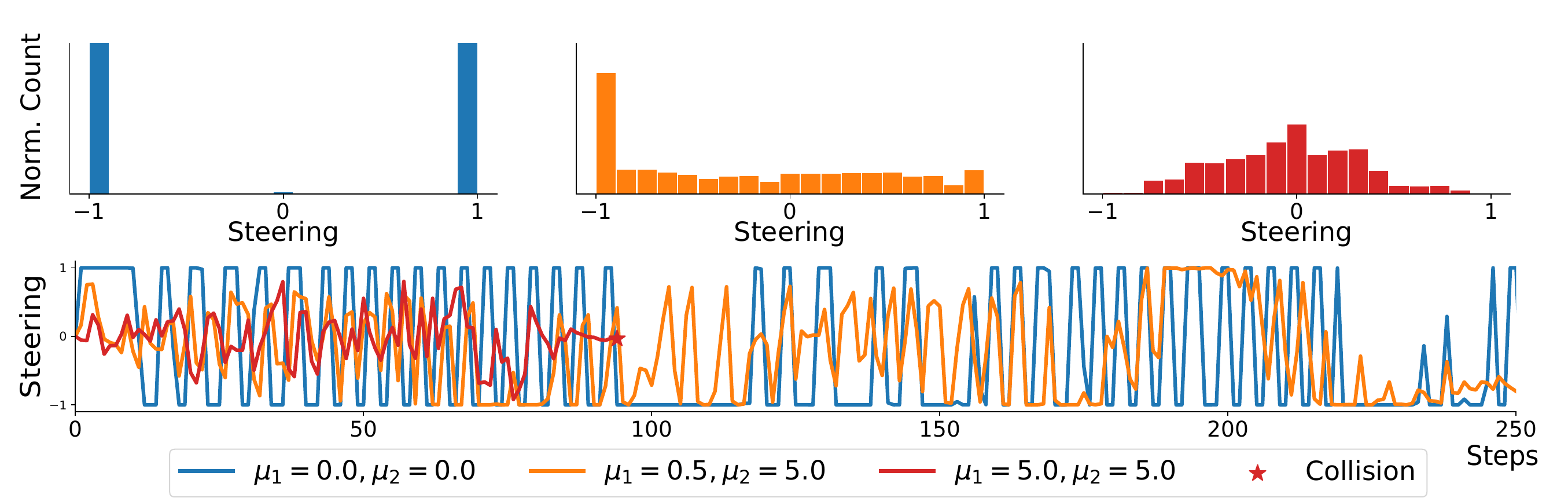}
    \caption{Impact of action regularization on TRT. The first row reports the steering command's distributions under various regularization weights. The second row reports the continuous command over a single simulation snapshot. n=5}
    \label{fig:smoothed}
\end{figure*}
Considering the learning methods shown in Figure~\ref{fig:learning_curves_and_transfer} (right, top row), 
all the algorithms can complete the simple track, COL.
However, only Dreamer can generalize the racing task and transfer the learned skills to other complex tracks.
Generally, policies trained in AUT achieve better performance as they can complete at least one lap in each other track.    
Conversely, even though TRT contains challenging turns, the trained agents perform poorly on unseen complex tracks (AUT). The reason might be that most turns in TRT have the same direction, and consequently, the trained policy cannot generalize to altered scenarios. 

Moreover, we observed that the Dreamer-Occupancy agent shows lower generalization skills compared to Dreamer-Distance. The learned policy results in a more aggressive strategy, presenting a lower lap-time as shown in Figure~\ref{fig:learning_curves_and_transfer}~(right, bottom row).
However, the car often gets dangerously close to the track walls. This results in unsafe behavior that increases the probability of collisions.
Comparing the performance of Dreamer with the adaptive FTG, we observe similar lap-times on all the tracks. 
However, the predictable behavior of FTG results in a more stable controller.
This is not surprising considering that FTG is a programmed algorithm, and its parameters have been carefully tuned to drive on the most challenging track, AUT.
In conclusion, reconstructing the occupancy area allows the agent \emph{overfit} to the track it was trained on. This speeds up learning, but hurts robustness to domain changes and thus worsens generalization.

\noindent \textbf{Influence of imagination horizon.} Figure \ref{fig:compare_horizon} shows that as we increase the imagination horizon for Dreamer, it performs better. 
While this increase helps generate long-term trajectories in latent space as training data, we observe that a further increase leads to a drop in performance in some of the experiments. We suspect that this drop is caused by compounding model errors for longer horizons.

\noindent \textbf{Action Regularization.}
Additionally, we observed that the driving behavior of the car resembles an on-off control law, resulting in unnecessarily curvy trajectories. While this type of behavior is expected for acceleration control~\cite{liberzon2011cov_and_optimal_control}, it is counterproductive to perform unnecessary steering commands.
We experimented with different regularization techniques for continuous control to minimize the steering effort and enforce temporal smoothness, similar to the approaches in \cite{mysore2021regularizing} and \cite{seyde2021bang}. 
The resulting policy showed smoother steering commands (see Figure~\ref{fig:smoothed}) but did not show superior performance compared to the one trained without regularization.
The objective for the regularized action model is defined as
$$
\mathbb{E}_{p_{\theta}, q_{\phi}} \left[ \sum_{t=0}^{H} \gamma^t (V_{\lambda}(s_t) - \mu_1 ||a_t||^2_2 - \mu_2 ||a_{t}-a_{t-1}||^2_2) \right], \label{eq:regularized_loss}
$$
where $V_{\lambda}(s_t)$ denotes the original value estimation term, $\mu_1$ and $\mu_2$ are the control effort and the temporal smoothing coefficients, respectively.

\noindent \textbf{Sim2Real transfer.} In this experiment, we evaluate Dreamer policies with respect to their sim2real transferability.
We test trained dreamers deployed on the car in a physical test track. A video\footnote[1]{The video can be viewed at: \url{https://youtu.be/8ofWVLArZJQ}
} demonstrating the driving performance of the Dreamer agent is supplied with the submission. It presents the Dreamer agent completing a full lap in TRT in the forward direction.
Then, to evaluate its generalization capabilities, we tested the agent in reversed direction.
We observe that even if the agent was not trained in the reversed direction and not directly on the TRT track, the agent can complete a successful lap. Finally, we placed two obstacles after the most challenging turn and ran the agent in this configuration. The Dreamer agents can complete the lap and securely avoid obstacles.

\section{Conclusions} 
\label{sec:conclusion}
We show that Dreamer, a model-based deep RL algorithm, outperforms several other model-free RL algorithms in simulation.
Furthermore, we empirically demonstrate  that Dreamer is able to successfully transfer the policy that it learned in simulation to a real-world test environment without the use of explicit domain randomization techniques.
Ultimately, we show how observation models and model horizon affect generalization and domain adaptation of learned policies and that model-based agents can enable robust autonomy in real-world settings.

\noindent \textbf{Why Dreamer?} Dreamer is a comparably sample-efficient, high-performance deep RL algorithm. Its learned state space model can not only be used for policy learning, but also for trajectory planning approaches~\cite{hafner2019learning}. 

\noindent \textbf{How does the sim2real gap influence Dreamer's performance?}
The discrepancy between simulation and reality is and remains one of the main challenges in RL. However, Dreamer's latent state-space model is robust enough to transfer the learned policy to the real world. This result is key for the deployment of RL algorithms in the real world.

\noindent \textbf{What are the limitations of this approach?}
We observed that our learned policies often resemble bang-bang control \cite{seyde2021bang}, which is often not desirable in real-world robotics applications. To avoid this, the objective function of the learning problem has to be carefully designed. Our experiments mitigated its emergence by regularizing the original actor's loss with action penalties to discourage sharp changes between extreme values. This method relaxed the bang-bang behavior observed in the obtained policies.
Furthermore, the lack of structure in the latent space of the world model makes it hard to interpret and is subject to ongoing research. All code, data and appendix are available at: \url{https://github.com/CPS-TUWien/racing_dreamer}.

\section*{Acknowledgments}
L.B. was supported by the Doctoral College Resilient Embedded Systems.
M.L. was supported in part by the ERC-2020-AdG 101020093 and the Austrian Science Fund (FWF) under grant Z211-N23 (Wittgenstein Award). R.H. and D.R. were supported by The Boeing Company and the Office of Naval Research (ONR) Grant N00014-18-1-2830. R.G. was partially supported by the Horizon-2020 ECSEL Project grant No. 783163 (iDev40) and A.B. by FFG Project ADEX.

\bibliographystyle{IEEEtran}
\bibliography{references}

\begin{thebibliography}{10}
\providecommand{\url}[1]{#1}
\csname url@samestyle\endcsname
\providecommand{\newblock}{\relax}
\providecommand{\bibinfo}[2]{#2}
\providecommand{\BIBentrySTDinterwordspacing}{\spaceskip=0pt\relax}
\providecommand{\BIBentryALTinterwordstretchfactor}{4}
\providecommand{\BIBentryALTinterwordspacing}{\spaceskip=\fontdimen2\font plus
\BIBentryALTinterwordstretchfactor\fontdimen3\font minus
  \fontdimen4\font\relax}
\providecommand{\BIBforeignlanguage}[2]{{%
\expandafter\ifx\csname l@#1\endcsname\relax
\typeout{** WARNING: IEEEtran.bst: No hyphenation pattern has been}%
\typeout{** loaded for the language `#1'. Using the pattern for}%
\typeout{** the default language instead.}%
\else
\language=\csname l@#1\endcsname
\fi
#2}}
\providecommand{\BIBdecl}{\relax}
\BIBdecl

\bibitem{sutton2018reinforcement}
R.~S. Sutton and A.~G. Barto, \emph{RL: An introduction}.\hskip 1em plus 0.5em
  minus 0.4em\relax MIT press, 2018.

\bibitem{hafner2019dream}
D.~Hafner, T.~Lillicrap, J.~Ba, and M.~Norouzi, ``Dream to control: Learning
  behaviors by latent imagination,'' \emph{arXiv:1912.01603}, 2019.

\bibitem{ha2018worldmodels}
\BIBentryALTinterwordspacing
D.~Ha and J.~Schmidhuber, ``Recurrent world models facilitate policy
  evolution,'' in \emph{Advances in Neural Information Processing Systems},
  S.~Bengio, H.~Wallach, H.~Larochelle, K.~Grauman, N.~Cesa-Bianchi, and
  R.~Garnett, Eds., vol.~31.\hskip 1em plus 0.5em minus 0.4em\relax Curran
  Associates, Inc., 2018. [Online]. Available:
  \url{https://proceedings.neurips.cc/paper/2018/file/2de5d16682c3c35007e4e92982f1a2ba-Paper.pdf}
\BIBentrySTDinterwordspacing

\bibitem{hafner2020mastering}
D.~Hafner, T.~P. Lillicrap, M.~Norouzi, and J.~Ba, ``Mastering atari with
  discrete world models,'' in \emph{International Conference on Learning
  Representations}, 2020.

\bibitem{hafner2019learning}
D.~Hafner, T.~Lillicrap, I.~Fischer, R.~Villegas, D.~Ha, H.~Lee, and
  J.~Davidson, ``Learning latent dynamics for planning from pixels,'' in
  \emph{Int. Conf. on Machine Learning}.\hskip 1em plus 0.5em minus 0.4em\relax
  PMLR, 2019, pp. 2555--2565.

\bibitem{haarnoja2018soft}
T.~Haarnoja, A.~Zhou, P.~Abbeel, and S.~Levine, ``Soft actor-critic: Off-policy
  maximum entropy deep reinforcement learning with a stochastic actor,'' in
  \emph{Int. Conf. on Machine Learning}, 2018, pp. 1861--1870.

\bibitem{barthmaron2018d4pg}
G.~Barth-Maron, M.~W. Hoffman, D.~Budden, W.~Dabney, D.~Horgan, D.~TB,
  A.~Muldal, N.~Heess, and T.~Lillicrap,
  ``\href{https://openreview.net/forum?id=SyZipzbCb}{Distributed Distributional
  Deterministic Policy Gradients},'' in \emph{International Conference on
  Learning Representations}, 2018.

\bibitem{schulman2017proximal}
J.~Schulman, F.~Wolski, P.~Dhariwal, A.~Radford, and O.~Klimov, ``Proximal
  policy optimization algorithms,'' 2017.

\bibitem{abdolmaleki2018mpo}
\BIBentryALTinterwordspacing
A.~Abdolmaleki, J.~T. Springenberg, Y.~Tassa, R.~Munos, N.~Heess, and
  M.~Riedmiller, ``Maximum a posteriori policy optimisation,'' in
  \emph{International Conference on Learning Representations}, 2018. [Online].
  Available: \url{https://openreview.net/forum?id=S1ANxQW0b}
\BIBentrySTDinterwordspacing

\bibitem{hasani2020natural}
R.~Hasani, M.~Lechner, A.~Amini, D.~Rus, and R.~Grosu, ``A natural lottery
  ticket winner: Reinforcement learning with ordinary neural circuits,'' in
  \emph{International Conference on Machine Learning}.\hskip 1em plus 0.5em
  minus 0.4em\relax PMLR, 2020, pp. 4082--4093.

\bibitem{yang2018sampleeff}
Y.~Yu, ``Towards sample efficient reinforcement learning,'' in
  \emph{Proceedings of the 27th International Joint Conference on Artificial
  Intelligence}, ser. IJCAI'18.\hskip 1em plus 0.5em minus 0.4em\relax AAAI
  Press, 2018, p. 5739–5743.

\bibitem{mnih2013dqn}
V.~Mnih, K.~Kavukcuoglu, D.~Silver, A.~Graves, I.~Antonoglou, D.~Wierstra, and
  M.~Riedmiller, ``\href{https://arxiv.org/abs/1312.5602}{Playing atari with
  deep reinforcement learning},'' \emph{arXiv preprint arXiv:1312.5602}, 2013.

\bibitem{Zhu2020robotics}
\BIBentryALTinterwordspacing
H.~Zhu, J.~Yu, A.~Gupta, D.~Shah, K.~Hartikainen, A.~Singh, V.~Kumar, and
  S.~Levine, ``The ingredients of real world robotic reinforcement learning,''
  in \emph{International Conference on Learning Representations}, 2020.
  [Online]. Available: \url{https://openreview.net/forum?id=rJe2syrtvS}
\BIBentrySTDinterwordspacing

\bibitem{sim2real_survey_2020}
W.~{Zhao}, J.~P. {Queralta}, and T.~{Westerlund},
  ``\href{https://ieeexplore.ieee.org/document/9308468}{Sim-to-Real Transfer in
  Deep RL for Robotics: a Survey},'' in \emph{IEEE Symposium Series on
  Computational Intelligence (SSCI)}, 2020, pp. 737--744.

\bibitem{peng2018sim2real}
\BIBentryALTinterwordspacing
X.~B. Peng, M.~Andrychowicz, W.~Zaremba, and P.~Abbeel, ``Sim-to-real transfer
  of robotic control with dynamics randomization,'' \emph{2018 IEEE
  International Conference on Robotics and Automation (ICRA)}, May 2018.
  [Online]. Available: \url{http://dx.doi.org/10.1109/ICRA.2018.8460528}
\BIBentrySTDinterwordspacing

\bibitem{sim2real_quadruped_robot_2018}
J.~Tan, T.~Zhang, E.~Coumans, A.~Iscen, Y.~Bai, D.~Hafner, S.~Bohez, and
  V.~Vanhoucke,
  ``\href{http://www.roboticsproceedings.org/rss14/p10.pdf}{Sim-to-Real:
  Learning Agile Locomotion For Quadruped Robots},'' in \emph{Proceedings of
  Robotics: Science and Systems}, Pittsburgh, Pennsylvania, June 2018.

\bibitem{kaspar2020sim2real}
M.~{Kaspar}, J.~D. {Muñoz Osorio}, and J.~{Bock},
  ``\href{https://ras.papercept.net/proceedings/IROS20/1205.pdf}{Sim2Real
  Transfer for Reinforcement Learning without Dynamics Randomization},'' in
  \emph{2020 IEEE/RSJ International Conference on Intelligent Robots and
  Systems (IROS)}, 2020, pp. 4383--4388.

\bibitem{Haarnoja2019learningToWalk}
T.~Haarnoja, S.~Ha, A.~Zhou, J.~Tan, G.~Tucker, and S.~Levine, ``Learning to
  walk via deep reinforcement learning,'' in \emph{Proceedings of Robotics:
  Science and Systems}, FreiburgimBreisgau, Germany, June 2019.

\bibitem{Levine_EndToEnd_withoutRewardEngineering}
A.~Singh, L.~Yang, C.~Finn, and S.~Levine,
  ``\href{http://www.roboticsproceedings.org/rss15/p73.pdf}{End-To-End Robotic
  Reinforcement Learning without Reward Engineering},'' in \emph{Proceedings of
  Robotics: Science and Systems}, 2019.

\bibitem{sutton1991dyna}
\BIBentryALTinterwordspacing
R.~S. Sutton, ``Dyna, an integrated architecture for learning, planning, and
  reacting,'' \emph{SIGART Bull.}, vol.~2, no.~4, p. 160–163, Jul. 1991.
  [Online]. Available: \url{https://doi.org/10.1145/122344.122377}
\BIBentrySTDinterwordspacing

\bibitem{deisenroth2011pilco}
M.~P. Deisenroth and C.~E. Rasmussen,
  ``\href{https://dl.acm.org/doi/10.5555/3104482.3104541}{PILCO: A Model-Based
  and Data-Efficient Approach to Policy Search},'' in \emph{Proceedings of the
  28th International Conference on Machine Learning}, ser. ICML'11.\hskip 1em
  plus 0.5em minus 0.4em\relax Madison, WI, USA: Omnipress, 2011, p. 465–472.

\bibitem{schwarting2021deep}
W.~Schwarting, T.~Seyde, I.~Gilitschenski, L.~Liebenwein, R.~Sander,
  S.~Karaman, and D.~Rus, ``Deep latent competition: Learning to race using
  visual control policies in latent space,'' \emph{arXiv preprint
  arXiv:2102.09812}, 2021.

\bibitem{kabzan2019amz}
J.~Kabzan, M.~de~la Iglesia~Valls, V.~Reijgwart, H.~F.~C. Hendrikx, C.~Ehmke,
  M.~Prajapat, A.~Bühler, N.~Gosala, M.~Gupta, R.~Sivanesan, A.~Dhall,
  E.~Chisari, N.~Karnchanachari, S.~Brits, M.~Dangel, I.~Sa, R.~Dubé,
  A.~Gawel, M.~Pfeiffer, A.~Liniger, J.~Lygeros, and R.~Siegwart, ``Amz
  driverless: The full autonomous racing system,'' 2019.

\bibitem{kabzan2019model}
J.~Kabzan, L.~Hewing, A.~Liniger, and M.~N. Zeilinger,
  ``\BIBforeignlanguage{en}{Learning-based model predictive control for
  autonomous racing},'' \emph{\BIBforeignlanguage{en}{IEEE Robotics and
  Automation Letters}}, vol.~4, no.~4, pp. 3363 -- 3370, 2019-10.

\bibitem{andresen2020mappingplanning_autonomousracing}
L.~{Andresen}, A.~{Brandemuehl}, A.~{Honger}, B.~{Kuan}, N.~{Vödisch},
  H.~{Blum}, V.~{Reijgwart}, L.~{Bernreiter}, L.~{Schaupp}, J.~J. {Chung},
  M.~{Burki}, M.~R. {Oswald}, R.~{Siegwart}, and A.~{Gawel},
  ``\href{https://ras.papercept.net/proceedings/IROS20/0988.pdf}{Accurate
  Mapping and Planning for Autonomous Racing},'' in \emph{2020 IEEE/RSJ Int.
  Conference on Intelligent Robots and Systems (IROS)}, 2020, pp. 4743--4749.

\bibitem{velenis1528936}
E.~{Velenis} and P.~{Tsiotras},
  ``\href{https://ieeexplore.ieee.org/document/1528936}{Minimum Time vs Maximum
  Exit Velocity Path Optimization During Cornering},'' in \emph{Proceedings of
  the IEEE International Symposium on Industrial Electronics, 2005. ISIE
  2005.}, vol.~1, 2005, pp. 355--360.

\bibitem{rucco7047758}
A.~{Rucco}, G.~{Notarstefano}, and J.~{Hauser},
  ``\href{https://ieeexplore.ieee.org/document/7047758}{An Efficient
  Minimum-Time Trajectory Generation Strategy for Two-Track Car Vehicles},''
  \emph{IEEE Trans on Control Systems Tech}, vol.~23, no.~4, pp. 1505--1519,
  2015.

\bibitem{timings2013minimum}
J.~P. Timings and D.~J. Cole,
  ``\href{https://asmedigitalcollection.asme.org/dynamicsystems/article/135/3/031015/370331/Minimum-Maneuver-Time-Calculation-Using-Convex}{Minimum
  maneuver time calculation using convex optimization},'' \emph{Journal of
  Dynamic Systems, Measurement, and Control}, vol. 135, no.~3, 2013.

\bibitem{vazquez2020hierarchical_motion_planning}
J.~L. {Vázquez}, M.~{Brühlmeier}, A.~{Liniger}, A.~{Rupenyan}, and
  J.~{Lygeros}, ``Optimization-based hierarchical motion planning for
  autonomous racing,'' in \emph{2020 IEEE/RSJ International Conference on
  Intelligent Robots and Systems (IROS)}, 2020, pp. 2397--2403.

\bibitem{jaritz_EndToEndRaceDrivingWithDeepRL}
M.~{Jaritz}, R.~{de Charette}, M.~{Toromanoff}, E.~{Perot}, and
  F.~{Nashashibi}, ``End-to-end race driving with deep reinforcement
  learning,'' in \emph{2018 IEEE International Conference on Robotics and
  Automation (ICRA)}, 2018, pp. 2070--2075.

\bibitem{riedmiller_LearningToDriveARealCarIn20Minutes}
M.~{Riedmiller}, M.~{Montemerlo}, and H.~{Dahlkamp}, ``Learning to drive a real
  car in 20 minutes,'' in \emph{2007 Frontiers in the Convergence of Bioscience
  and Information Technologies}, 2007, pp. 645--650.

\bibitem{kendall_LearningToDriveInADay}
A.~{Kendall}, J.~{Hawke}, D.~{Janz}, P.~{Mazur}, D.~{Reda}, J.~{Allen},
  V.~{Lam}, A.~{Bewley}, and A.~{Shah}, ``Learning to drive in a day,'' in
  \emph{2019 International Conference on Robotics and Automation (ICRA)}, 2019,
  pp. 8248--8254.

\bibitem{fuchs2020super}
\BIBentryALTinterwordspacing
F.~Fuchs, Y.~Song, E.~Kaufmann, D.~Scaramuzza, and P.~Duerr,
  ``\href{https://arxiv.org/abs/2008.07971}{Super-Human Performance in Gran
  Turismo Sport Using Deep Reinforcement Learning},'' \emph{arXiv preprint
  arXiv:2008.07971}, 2020. [Online]. Available:
  \url{https://arxiv.org/abs/2008.07971}
\BIBentrySTDinterwordspacing

\bibitem{bellegarda2020mpc_drl_trajplanning}
G.~{Bellegarda} and K.~{Byl},
  ``\href{https://ras.papercept.net/proceedings/IROS20/2178.pdf}{An Online
  Training Method for Augmenting MPC with Deep Reinforcement Learning},'' in
  \emph{2020 IEEE/RSJ Int. Conf. on Intelligent Robots and Systems (IROS)},
  2020, pp. 5453--5459.

\bibitem{williams2017mpc}
G.~{Williams}, N.~{Wagener}, B.~{Goldfain}, P.~{Drews}, J.~M. {Rehg},
  B.~{Boots}, and E.~A. {Theodorou}, ``Information theoretic mpc for
  model-based reinforcement learning,'' in \emph{2017 IEEE International
  Conference on Robotics and Automation (ICRA)}, 2017, pp. 1714--1721.

\bibitem{schaal1999imitation}
S.~Schaal, ``Is imitation learning the route to humanoid robots?'' \emph{Trends
  in cognitive sciences}, vol.~3, no.~6, pp. 233--242, 1999.

\bibitem{ross2011reduction}
S.~Ross, G.~Gordon, and D.~Bagnell, ``A reduction of imitation learning and
  structured prediction to no-regret online learning,'' in \emph{Proceedings of
  the fourteenth international conference on artificial intelligence and
  statistics}.\hskip 1em plus 0.5em minus 0.4em\relax JMLR Workshop and
  Conference Proceedings, 2011, pp. 627--635.

\bibitem{vorbach2021causal}
C.~Vorbach, R.~Hasani, A.~Amini, M.~Lechner, and D.~Rus, ``Causal navigation by
  continuous-time neural networks,'' \emph{Advances in Neural Information
  Processing Systems}, vol.~34, 2021.

\bibitem{lechner2020gershgorin}
M.~Lechner, R.~Hasani, D.~Rus, and R.~Grosu, ``Gershgorin loss stabilizes the
  recurrent neural network compartment of an end-to-end robot learning
  scheme,'' in \emph{2020 IEEE International Conference on Robotics and
  Automation (ICRA)}.\hskip 1em plus 0.5em minus 0.4em\relax IEEE, 2020, pp.
  5446--5452.

\bibitem{ng2000algorithms}
A.~Y. Ng, S.~J. Russell \emph{et~al.}, ``Algorithms for inverse reinforcement
  learning.'' in \emph{Icml}, vol.~1, 2000, p.~2.

\bibitem{wu2019imitation}
Y.-H. Wu, N.~Charoenphakdee, H.~Bao, V.~Tangkaratt, and M.~Sugiyama,
  ``Imitation learning from imperfect demonstration,'' in \emph{International
  Conference on Machine Learning}.\hskip 1em plus 0.5em minus 0.4em\relax PMLR,
  2019, pp. 6818--6827.

\bibitem{sun2019adversarial}
M.~Sun and X.~Ma, ``Adversarial imitation learning from incomplete
  demonstrations,'' \emph{arXiv preprint arXiv:1905.12310}, 2019.

\bibitem{lechner2021adversarial}
M.~Lechner, R.~Hasani, R.~Grosu, D.~Rus, and T.~A. Henzinger, ``Adversarial
  training is not ready for robot learning,'' in \emph{2021 IEEE International
  Conference on Robotics and Automation (ICRA)}.\hskip 1em plus 0.5em minus
  0.4em\relax IEEE, 2021, pp. 4140--4147.

\bibitem{lechner2020neural}
M.~Lechner, R.~Hasani, A.~Amini, T.~A. Henzinger, D.~Rus, and R.~Grosu,
  ``Neural circuit policies enabling auditable autonomy,'' \emph{Nature Machine
  Intelligence}, vol.~2, no.~10, pp. 642--652, 2020.

\bibitem{lechner2019designing}
M.~Lechner, R.~Hasani, M.~Zimmer, T.~A. Henzinger, and R.~Grosu, ``Designing
  worm-inspired neural networks for interpretable robotic control,'' in
  \emph{2019 International Conference on Robotics and Automation (ICRA)}.\hskip
  1em plus 0.5em minus 0.4em\relax IEEE, 2019, pp. 87--94.

\bibitem{amini2020learning}
A.~Amini, I.~Gilitschenski, J.~Phillips, J.~Moseyko, R.~Banerjee, S.~Karaman,
  and D.~Rus, ``Learning robust control policies for end-to-end autonomous
  driving from data-driven simulation,'' \emph{IEEE Robotics and Automation
  Letters}, vol.~5, no.~2, pp. 1143--1150, 2020.

\bibitem{pybullet}
E.~Coumans and Y.~Bai, ``Pybullet, a python module for physics simulation for
  games, robotics and machine learning,'' \url{http://pybullet.org},
  2016--2019.

\bibitem{babu2020f1tenth}
V.~S. Babu and M.~Behl, ``f1tenth. dev-an open-source ros based f1/10
  autonomous racing simulator,'' in \emph{16th Int. Conf. on Automation Science
  and Engineering (CASE)}.\hskip 1em plus 0.5em minus 0.4em\relax IEEE, 2020,
  pp. 1614--1620.

\bibitem{o2020tunercar}
M.~O’Kelly, H.~Zheng, A.~Jain, J.~Auckley, K.~Luong, and R.~Mangharam,
  ``Tunercar: A superoptimization toolchain for autonomous racing,'' in
  \emph{2020 IEEE International Conference on Robotics and Automation
  (ICRA)}.\hskip 1em plus 0.5em minus 0.4em\relax IEEE, 2020, pp. 5356--5362.

\bibitem{ros}
\BIBentryALTinterwordspacing
{Stanford Artificial Intelligence Laboratory et al.}, ``Robotic operating
  system.'' [Online]. Available: \url{https://www.ros.org}
\BIBentrySTDinterwordspacing

\bibitem{BRAGHIN20081503}
\BIBentryALTinterwordspacing
F.~Braghin, F.~Cheli, S.~Melzi, and E.~Sabbioni, ``Race driver model,''
  \emph{Computers \& Structures}, vol.~86, no.~13, pp. 1503--1516, 2008,
  structural Optimization. [Online]. Available:
  \url{https://www.sciencedirect.com/science/article/pii/S0045794908000163}
\BIBentrySTDinterwordspacing

\bibitem{Lamiraux2001Smoothmotionplanning}
F.~{Lamiraux} and J.~. {Lammond}, ``Smooth motion planning for car-like
  vehicles,'' \emph{IEEE Transactions on Robotics and Automation}, vol.~17,
  no.~4, pp. 498--501, 2001.

\bibitem{hochreiter1997long}
S.~Hochreiter and J.~Schmidhuber, ``Long short-term memory,'' \emph{Neural
  computation}, vol.~9, no.~8, pp. 1735--1780, 1997.

\bibitem{hasani2021liquid}
R.~Hasani, M.~Lechner, A.~Amini, D.~Rus, and R.~Grosu, ``Liquid time-constant
  networks,'' in \emph{Proceedings of the AAAI Conference on Artificial
  Intelligence}, vol.~35, no.~9, 2021, pp. 7657--7666.

\bibitem{hasani2019response}
R.~Hasani, A.~Amini, M.~Lechner, F.~Naser, R.~Grosu, and D.~Rus, ``Response
  characterization for auditing cell dynamics in long short-term memory
  networks,'' in \emph{2019 International Joint Conference on Neural Networks
  (IJCNN)}.\hskip 1em plus 0.5em minus 0.4em\relax IEEE, 2019, pp. 1--8.

\bibitem{lechner2020learning}
M.~Lechner and R.~Hasani, ``Learning long-term dependencies in
  irregularly-sampled time series,'' \emph{arXiv preprint arXiv:2006.04418},
  2020.

\bibitem{hasani2021closed}
R.~Hasani, M.~Lechner, A.~Amini, L.~Liebenwein, M.~Tschaikowski, G.~Teschl, and
  D.~Rus, ``Closed-form continuous-depth models,'' \emph{arXiv preprint
  arXiv:2106.13898}, 2021.

\bibitem{akiba2019optuna}
T.~Akiba, S.~Sano, T.~Yanase, T.~Ohta, and M.~Koyama, ``Optuna: A
  next-generation hyperparameter optimization framework,'' 2019.

\bibitem{ftg2012}
\BIBentryALTinterwordspacing
V.~Sezer and M.~Gokasan, ``A novel obstacle avoidance algorithm: “follow the
  gap method”,'' \emph{Robotics and Autonomous Systems}, vol.~60, no.~9, pp.
  1123--1134, 2012. [Online]. Available:
  \url{https://www.sciencedirect.com/science/article/pii/S0921889012000838}
\BIBentrySTDinterwordspacing

\bibitem{liberzon2011cov_and_optimal_control}
D.~Liberzon, \emph{Calculus of Variations and Optimal Control Theory: A Concise
  Introduction}.\hskip 1em plus 0.5em minus 0.4em\relax USA: Princeton
  University Press, 2011.

\bibitem{mysore2021regularizing}
S.~Mysore, B.~Mabsout, R.~Mancuso, and K.~Saenko, ``Regularizing action
  policies for smooth control with reinforcement learning,'' 2021.

\bibitem{seyde2021bang}
T.~Seyde, I.~Gilitschenski, W.~Schwarting, B.~Stellato, M.~Riedmiller,
  M.~Wulfmeier, and D.~Rus, ``Is bang-bang control all you need? solving
  continuous control with bernoulli policies,'' \emph{Advances in Neural
  Information Processing Systems}, vol.~34, 2021.

\end{thebibliography}

\end{document}